\documentclass{article}
\usepackage{spconf,amsmath,graphicx}
\usepackage{multicol}
\usepackage{lipsum}
\usepackage{subfigure}
\usepackage[justification=centering]{caption}
\usepackage{booktabs}

\usepackage{enumitem}

\title{Optical Character Recognition (OCR) for Telugu:  \\ Database, Algorithm and Application}
\name{Konkimalla Chandra Prakash, Y. M. Srikar, Gayam Trishal, Souraj Mandal, Sumohana S. Channappayya }
\address{Indian Institute of Technology Hyderabad, Kandi - 502285, Telangana, India}

\begin{document}
%
\maketitle
\begin{abstract}
Telugu is a Dravidian language spoken by more than 80 million people worldwide. The optical character recognition (OCR) of the Telugu script has wide ranging applications including education, health-care, administration etc. The beautiful Telugu script however is very different from Germanic scripts like English and German. This makes the use of transfer learning of Germanic OCR solutions to Telugu a non-trivial task. To address the challenge of OCR for Telugu, we make three contributions in this work: (i) a database of Telugu characters, (ii) a deep learning based OCR algorithm, and (iii) a client server solution for the online deployment of the algorithm. For the benefit of the Telugu people and the research community, our code has been made freely available at  https://gayamtrishal.github.io/OCR\_Telugu.github.io/.
\end{abstract}
\section{Introduction}
\label{sec:intro}

Telugu is the official language of the Indian states of Telangana and Andhra Pradesh. It ranks third by the number of native speakers in India, and fifteenth in the Ethnologue list of most-spoken languages worldwide \cite{wiki:xxx}. There are a large number of Telugu character shapes whose components are simple and compound characters from 16 vowels (called \emph{achus}) and 36 consonants (called \emph{hallus}). Optical Character Recognition (OCR) is the mechanical or electronic conversion of images of typed, handwritten or printed text into machine-encoded text. The availability of huge online collections of scanned Telugu documents in conjunction with applications in e-governance and healthcare justifies the necessity for an OCR system, but the complex script and grammar make the problem a challenging one.  

\begin{figure}[htbp]
\centering

\includegraphics[height = 2cm]{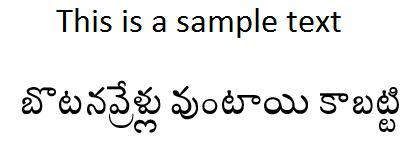}

  \caption{English text vs Telugu text.}
    \label{fig:eng_tel}
    \vspace{-1em}

\end{figure}

OCR for Indian languages is much more challenging than that of Germanic languages because of the huge number of combinations of the main characters, \emph{vattus}, and \emph{guninthas} (modifiers). Unlike Germanic languages, Telugu characters are round in shape, and seldom contain any horizontal or vertical lines.  In the English language, character segmentation can be easily done using connected components-like algorithms, as a majority of the characters are formed by a single stroke. In the Telugu script however, parts of the character extend both above and below the main characters and are also not joined to the main character as shown in \ref{fig:eng_tel}. This makes the use of histogram based segmentation methods (and transfer learning in general) difficult.  

The complexity of the problem at hand is huge because of the large number of output classes possible and the inter class variability. The absence of robust deep learning based OCR systems for Telugu has motivated us to build one. An OCR system has a huge impact in real life applications along with a word processor. In the literature, other attempts on Telugu OCR have neither shown results on large datasets \cite{jawahar2003bilingual} nor have considered all possible character and \emph{vattu} combinations 
which exist in the language. Here, we describe a novel end-to-end approach for Telugu OCR.

Besides data, classifier selection also has a significant impact on an OCR system. Before deep learning was used, feature learning was a critical step in the design of any classifier because feeding raw data would not lead to the targeted results. Therefore, classification is generally performed after the difficult process of appropriate feature selection that distinguishes classes. The advent of Convolutional Neural Networks (CNNs) has paved the way for automated feature learning. Also, the strong generalization capability of this multi-layered network has pushed the classification performance beyond human accuracy. Due to these reasons, we have used a CNN based classifier in our OCR system. 

We have addressed these challenges in Telugu OCR and summarize our contributions as follows:
	\begin{itemize}[noitemsep,topsep=0pt,parsep=0pt,partopsep=0pt]
    	\item We introduce the largest dataset for Telugu characters with 17387 categories and 560 samples per category.
        \item We propose a 2-CNN architecture that performs extremely well on our dataset.
        \item We have developed an android application for its deployment.

    \end{itemize}

The rest of the paper is structured as follows. Section \ref{sec:related} talks about previous works on OCR of Telugu and Kannada (a similar script). Section \ref{sec:method} briefly describes the methodology and novelties introduced in this paper. Sub-section \ref{ssec:dataset} talks about the proposed dataset. Sub-section \ref{ssec:classifier} presents the architectural details of the proposed CNN framework and the overall model for classification of characters. Sub-section \ref{ssec:preprocessing} gives an in-depth explanation of the prepossessing steps and segmentation algorithm. We finally present results in Section \ref{sec:results} and offer concluding remarks in Section \ref{sec:conclusions}.

\section{RELATED WORK}
\label{sec:related}

Optical character recognition (OCR) has been one of the most studied problems in pattern recognition. Until recently, feature engineering was the dominant approach that used features like Wavelet features, Gabor features, Circular features, Skeleton features etc; \cite{babu2014ocr} \cite{ramanathan2009robust} \cite{hu2002recognition} followed by a support vector machine (SVM) or boosting based classifiers. The recent and astounding success of CNNs in feature learning has motivated us to use them for Telugu character recognition.

The first reported work on OCR for Telugu can be dated back to 1977 by Rajasekharan and Deekshatulu \cite{rajasekaran1977recognition}  which used features that encode the curves that trace a letter, and compare this encoding with a set of predefined templates. It was able to identify 50 primitive features, and proposes a two-stage syntax-aided character recognition system. The first attempt to use neural networks was made by Sukhaswami et al., which trains multiple neural networks, pre-classifies an image based on its aspect ratio and feeds it to the corresponding network \cite{rao1995telugu}. It demonstrated the robustness of a Hopfield network for the purpose of recognition of noisy Telugu characters.  Later work on Telugu OCR primarily followed the featurization classification paradigm \cite{an2001system}.

The work by Jawahar et al., \cite{jawahar2003bilingual} describes a bilingual Hindi-Telugu OCR for documents containing Hindi and Telugu text. It is based on Principal Components Analysis (PCA) followed by support vector regression. They report an overall accuracy of 96.7\% over an independent test set. They perform character level segmentation offline using their data collecting tools. However, they have only considered 330 distinct classes. 

The work by Achanta and Hastie \cite{achanta2015telugu} on Telugu OCR using convolutional neural networks is also interesting. They used 50 fonts in four styles for training data each image of size $48 \times 48$. However, they did not consider all possible outputs (only 457 classes) of CNN. The work by Kunte and Samuel \cite{kunte2007ocr} on Kannada OCR employs a twp-stage classification system that is similar to our approach. They have first used wavelets for feature extraction and then two-stage multi-layer perceptrons for the task of classification. They have divided the characters into seperate sub classes but have not considered all possible combinations. 

Our proposed approach addresses some of the shortcomings in the literature and is described next.

\section{PROPOSED METHODOLOGY}
\label{sec:method}
The pipeline followed here is a classic one: skew correction -- word segmentation -- character segmentation -- recognition. Our paper introduces novelties in the dataset and classifier. Our pre-processing and segmentation techniques are minor modifications of existing techniques and are fine-tuned for the Telugu script as described in the following subsections. We describe the dataset next followed by a description of the classifier and the application.

\subsection{Dataset}
\label{ssec:dataset}
A major issue in the field of Telugu OCR
is a lack of large data repositories of Telugu characters that are needed for training deep neural networks. This could be attributed to the fact that most previous methods (as described in the previous Section) did not rely on deep learning techniques for feature extaction.

For e.g., the work by Pramod et al., \cite{sankar2010nearest} has 1000 words and on an average of 32 images per category. They used the most frequently occuring words in Telugu but were unable to cover all the words in the Telugu language. Later works were based on character level \cite{babu2014ocr} \cite{prasanth2007elastic} \cite{kunte2007ocr}. The dataset by Achanta and Hastie \cite{achanta2015telugu} has 460 classes and 160 samples per class which made up 76000 images. However, these works have not considered all the possible combinations of \emph{vattu} and \emph{guninthas}. To tackle this issue,  we propose a dataset which takes into consideration all possible combinations of \emph{vattu} and \emph{guninthas}. This is to ensure that the classification algorithms have a good set of training samples which in turn helps improve overall performance.
 
Each character has been augmented with 20 different fonts downloaded from \cite{fonts}. Using all the fonts for {\em{gutintham}} variants and 3 fonts for {\em{vattu}} variants, all possible \emph{vattu} and \emph{gunintham} forms of a character have been manually entered in Microsoft Word. We then changed the font size from 15 to 40 with a step size of 5 covering 6 different font sizes for all the variants of each character. We then took screen-shots of each page containing these characters and used our segmentation algorithm on them to get the individual characters. 

\begin{figure*}
\begin{multicols}{2}
    \centering
    \hbox{\hspace{2.5em}
    \subfigure[Architecture 1 for main character.]{\includegraphics[width=13cm]{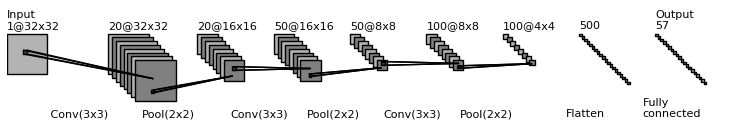}}
    }
\end{multicols}

\begin{multicols}{2}
    \centering
    \hbox{\hspace{2.5em}
    \subfigure[Architecture 2 for \emph{vattu}. ]{\includegraphics[width=13cm]{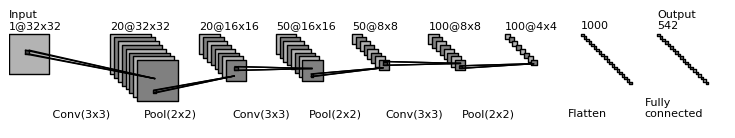}}
    }
\end{multicols}
\centering
\caption{Architecture of 1st CNN (main character) and 2nd CNN (\emph {vattu} and \emph {gunintham.})}
\label{fig:arc}
\vspace{-1em}
\end{figure*}

We have also introduced random rotations (angle in degrees: -6, -2, 2, 6), additive noise (variance $=$ 0.5 + $\frac{J}{10}$ * $\frac{2}{3}$ , $J$ $\in$ (0,5)) and random crops to simulate realistic conditions. We have then applied elastic deformations on the characters. The dataset has 17387 categories and nearly 560 samples per class. All the images are of size $32 \times 32$. There are 6,757,044 training samples, 972,309 validation samples and 1,934,190 test samples which add upto 1 million images (10 GB).  Our dataset is novel because unlike other datasets which only take into account the commonly occuring permutations of characters and \emph{vattu}s, we have spanned the entire Telugu alphabets and their corresponding \emph{vattu} and \emph{guninthas}.

\subsection{Classifier}
\label{ssec:classifier}
The performance of an OCR system depends hugely on the performance of its classifier. Previous works \cite{singh2010ocr} on Telugu OCR have done the character level segmentation based on histograms along the $x$ and $y$ directions. Assuming that the histogram method for segmentation would work perfectly, they have used an SVM based classifiers for character classification. However, we have observed that in real scenarios, the histogram method fails to properly segment out the \emph{vattu} and the main character together. It also fails when the characters are rotated or if they share common region when projected on $x$-axis or $y$-axis.

Inspired by the success of deep neural networks for feature learning, we have explored CNNs to classify the characters and proposed a new architecture for the same. A CNN is a type of feed-forward neural network or a sequence of multiple layers which is inspired by biological processes. It eliminates the dependency on hand-crafted features and directly learns useful features from the training data itself. It is a combination of both a feature extractor and a classifier and mainly consists of convolutional (weight-sharing), pooling and fully connected layers.

\begin{figure}[htbp]
\centering
\includegraphics[height = 2.5cm]{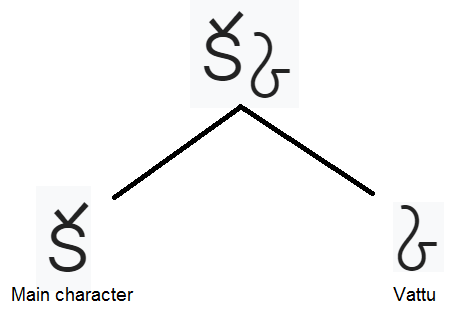}
  \caption{Main character and \emph{vattu}.}
    \label{fig:main_char}
    \vspace{-1em}
\end{figure}

In general, a Telugu character consists of two main components - the main character and the \emph{vattu}/\emph{gunintham} as shown in Figure \ref{fig:main_char}. Using a single CNN would be futile because of the huge number of classes arising from various permutations of the main character, \emph{vattu} and \emph{gunintham}. Therefore, we have used a 2 CNN architecture for classifying the character. The first CNN is used for identifying the main character and the second CNN for identifying the \emph{vattu} and/or \emph{gunintam} present along with the main character. The architectures for both the CNNs is shown in Figure \ref{fig:arc}.

\subsection{Pre-Processing and Segmentation}
\label{ssec:preprocessing}
Pre-processing consists of skew correction in which tilt in the image is adjusted and binarization after which the image is binarized and segmented into individual words. This is followed by character level segmentation and classification.

Our segmentation algorithm assumes that there is no skew in the image. This makes skew correction a matter of utmost importance. We have used a straight line Hough transform based technique for correcting skew that can detect and correct skew upto 90 degrees.
We used a modified version of Otsu's thresholding \cite{otsu1979threshold} for our binarization. We used morphological closing algorithm for noise removal. We then computed the logical OR between the denoised image and that of the Otsu's thresholding result and applied mode based threshold on it.

For application to Telugu characters, we modified the MSER method \cite{nister2008linear} to take into consideration \emph{dheergas} and \emph{vattu}s. In order to eliminate the possibility of \emph{dheergam} and \emph{vattu} being segmented separately we merged the nearby characters into one word by dilating the output of MSER.

We used the connected components algorithm for character level segmentation. After binarization of the word, we apply the algorithm to separate all the characters as components (groups of binary pixels). In this process, minor blobs are removed from the components. In some cases, \emph{vattu}s are not connected with the main/base character. So, for connecting the base character with its \emph{vattu}, we measured the overlapping distance in horizontal and vertical direction and grouped them together. 

\subsection{Mobile Application}
To facilitate usability, we have developed an Android app that deploys the proposed OCR solution. The app does online image to text conversion with Industry standard (MVP Architecture) which works on any Android (4.4+) device. We used client-server based communication where the client (App user) requests the server with an image and the server responds to the app with a html file. The theme is specifically made keeping in mind that old age or low vision people can use it easily. The App uses camera or gallery for images. This App will also be made publicly available.

\section{Results and Discussion}
\label{sec:results}
We now present the experimental details and the results of our proposed algorithm.
As presented earlier, a total of 6,757,044 samples were used for training the network. The performance was validated using 972,309 samples. We used a batch size of 500 because of the large training data size. Initially, we trained our network using a SGD + momentum optimizer. Even after 70-80 epochs, the accuracy was not satisfactory (80\%). By using the Adam optimizer, we were able to attain much higher accuracy within 30-40 epochs. We halted the training process when there is no increase in validation accuracy for a few epochs (5). Our model was trained on GTX 1060 with 16GB RAM. 

We would like to note that in addition to the CNN architectures that have been proposed in Fig. \ref{fig:arc}, standard CNN architectures defined in Cifar \cite{krizhevsky2010convolutional} and Lenet \cite{lecun2015lenet} were also trained using the same approach descirbed above. This was done primarily for comparative analysis as described next.


\begin{table}[htbp]
\centering
\caption{CNN accuracies for Character Classification.}
\label{tab:char}
\begin{tabular}{|c|c|c|}
    \hline
    \textbf{Network} & \textbf{Architecture} & \textbf{Accuracy} \\ \hline
    MC Cifar       & CRPC32-CRPC32-CRPC64-D360 & 98.60    \\ \hline
    MC Lenet       & CRPL20-CRPL50-D500 & 98.62    \\ \hline
    {\em{TCCNN-S}}       & {\em{CRP25-CRP20-DD256}} & {\em{97.95}}    \\ \hline
	{\em{TCCNN-L}}        & {\em{CRP20-CRP50-CRP100-DD500}}  & {\em{98.74}}    \\ \hline
    \end{tabular}
\end{table}

\begin{table}[htbp]

\centering
\caption{CNN accuracies for \emph{Vattu}.}
\label{tab:vattu}
\begin{tabular}{|c|c|c|}
    \hline
    \textbf{Network} & \textbf{Architecture} & \textbf{Accuracy} \\ \hline
    MV Cifar        & CRPC32-CRPC32-CRPC64-D500 & 95.46    \\ \hline
    MV Lenet        & CRPL20-CRPL50-D500 & 95.59    \\ \hline
    {\em{TVCNN-S}}         & {\em{CRP25-CRP20-DD256}} & {\em{94.32}}    \\ \hline
    {\em{TVCNN-L}}         & {\em{CRP20-CRP50-CRP100-DD1000}}  & {\em{96.09}}    \\ \hline

\end{tabular}
\end{table}

\subsection{Table descriptions}
Tables \ref{tab:char} and \ref{tab:vattu} show the accuracy of various CNN architectures on our testing data after the CNN was trained on the proposed dataset. The abbreviations in the tables are explained below.
\begin{itemize}[noitemsep,topsep=0pt,parsep=0pt,partopsep=0pt]
    	\item CRP (n) - Convolution (3x3, n filters), Relu, Pool(2x2)
        \item CRPC (n) - Convolution (3x3, n filters), Relu, Pool(3x3)
        \item D (n) - Dense layer of n nodes.
        \item DD (n) - Dropout and Dense layer of n nodes.
        \item TCCNN-L/S - Telugu Character CNN Large/Small
        \item TVCNN-L/S - Telugu Vattu CNN Large/Small
        \item MC/MV Cifar - Modified Character/Vattu Cifar
        \item MC/MV Lenet - Modified Character/Vattu Lenet
\end{itemize}
    
The last layers of the Cifar \cite{krizhevsky2010convolutional} and Lenet \cite{lecun2015lenet} architectures have been modified according to the number of outputs of the main character and \emph{vattu}. We have introduced two different architectures, each having two CNNs -- one for the main character and one for the \emph{vattu}. TCCNN-S and TVCNN-S are smaller architectures which are faster than the others but with slightly lower accuracy. TCCNN-L and TVCNN-L achieve better accuracy than both the Cifar and Lenet architectures. Even though the improvement is small, it is signficant due to the large size of our dataset. This improvement could be explained by the fact that the proposed architectures in Figs. \ref{fig:arc} are tuned for characters and {\em{vattu}} individually. Further, the architecture does not reduce the resolution of the image patch as much as the other architectures thereby helping with classification of subtle shapes in the Telugu character set. 

We have not used very deep models like VGG \cite{simonyan2014very} and Resnet \cite{he2016deep} because they are trained with input images of size $224 \times 224$. In our case however, the images are of size $32 \times 32$.. We couldn't compare with other works on Telugu OCR because there are very few which have character level segmentation and use a deep learning based approach for classification. The work by Achanta and Hastie \cite{achanta2015telugu} is the closest one to ours but has $48 \times 48$ images. On the other hand, our images are $32 \times 32$. CNN's structure varies with the image size, so such comparison would be futile. Further, their classes also differ. Hence, we have compared with standard CNN architectures on our dataset.
\section {Conclusion and Future Work}
\label{sec:conclusions}
We have presented a solution for Telugu OCR that includes a database, algorithm and an application. We have spanned the entire Telugu language while creating the dataset, so there isn't any further possibility of increase in data. The segmentation algorithm can be improved so that every character is segmented together with its \emph{vattu} and \emph{gunintham}. Network accuracy can be further improved to make the classifier better. This proposed work can be further extended to other languages with the scope of having a common OCR system for all the languages of India.
\bibliographystyle{IEEEbib}
\bibliography{ref}
\end{document}